\documentclass{article}

\usepackage[utf8]{inputenc}
\usepackage[english]{babel}
\usepackage[ruled,vlined]{algorithm2e}
\usepackage{spconf}
\usepackage{amsmath}
\usepackage{amssymb}
\usepackage{hyperref}
\usepackage{graphicx}
\usepackage{tikz}
\usepackage{xcolor}
\usepackage{color}
\usepackage{ifthen}
\usepackage{booktabs}
\usepackage{subcaption}
\usepackage{wrapfig}

\usetikzlibrary{arrows,positioning}
\usetikzlibrary{shapes,decorations}
\usetikzlibrary{calc}
\usetikzlibrary{decorations.pathreplacing}
\usetikzlibrary{decorations.text}
\usetikzlibrary{shapes.callouts}
\usetikzlibrary{positioning}
\usetikzlibrary{quotes}
\usetikzlibrary{arrows.meta}

\makeatletter
\newcommand{\subalign}[1]{%
    \vcenter{%
        \Let@ \restore@math@cr \default@tag
        \baselineskip\fontdimen10 \scriptfont\tw@
        \advance\baselineskip\fontdimen12 \scriptfont\tw@
        \lineskip\thr@@\fontdimen8 \scriptfont\thr@@
        \lineskiplimit\lineskip
        \ialign{\hfil$\m@th\scriptstyle##$&$\m@th\scriptstyle{}##$\hfil\crcr
        #1\crcr
        }%
    }%
}
\makeatother

\addto\extrasenglish{

}
\newcommand{\argmax}{\operatornamewithlimits{arg\,max}}

\definecolor{rwthblue}{RGB}{0, 84, 159}
\definecolor{rwthlightblue}{RGB}{64,127,183}
\definecolor{rwthmagenta}{RGB}{227,0,102}
\definecolor{rwthyellow}{RGB}{255,237,0}
\definecolor{rwthpetrol}{RGB}{0, 97, 101}
\definecolor{rwthturquoise}{RGB}{0, 152, 161}
\definecolor{rwthgreen}{RGB}{87, 171, 39}
\definecolor{rwthmaygreen}{RGB}{189, 205, 0}
\definecolor{rwthorange}{RGB}{246,168,0}
\definecolor{rwthred}{RGB}{204,7,30}
\definecolor{rwthbordeaux}{RGB}{161,16,53}
\definecolor{rwthviolet}{RGB}{97,33,88}
\definecolor{rwthpurple}{RGB}{122,111,172}

\newif\ifRenderfigures
\newif\ifRendertables
\Renderfigurestrue
\Rendertablestrue

\title{Efficient Sequence Training of Attention Models using \\Approximative Recombination}

\name{
    Nils-Philipp Wynands\textsuperscript{1},
    Wilfried Michel\textsuperscript{1,2},
    Jan Rosendahl\textsuperscript{1},
    Ralf Schlüter\textsuperscript{1,2},
    Hermann Ney\textsuperscript{1,2}
}
\address{
    \textsuperscript{1}Human Language Technology and Pattern Recognition, Computer Science Department,\\
    RWTH Aachen University, 52062 Aachen, Germany,\\
    \textsuperscript{2}AppTek GmbH, 52062 Aachen, Germany\\
    \small\texttt{nils-philipp.wynands@rwth-aachen.de, \{michel,rosendahl,schlueter,ney\}@cs.rwth-aachen.de}
}

\begin{document}
    %
    \maketitle
    \begin{abstract}
        Sequence discriminative training is a great tool to improve the performance of an automatic speech recognition
        system. It does, however, necessitate a sum over all possible word sequences, which is intractable to compute in
        practice. Current state-of-the-art systems with unlimited label context circumvent this problem by limiting the
        summation to an n-best list of relevant competing hypotheses obtained from beam search.

        This work proposes to perform (approximative) recombinations of hypotheses during beam search, if they
        share a common local history.
        The error that is incurred by the approximation is analyzed and it is shown that using this technique the
        effective beam size can be increased by several orders of magnitude without significantly increasing the
        computational requirements.
        Lastly, it is shown that this technique can be used to effectively perform sequence discriminative training for
        attention-based encoder-decoder acoustic models on the LibriSpeech task.
    \end{abstract}

    \section{Introduction}
    \label{sec:intro}

    State-of-the-art approaches to \textit{automatic speech recognition} (ASR) make use of \textit{acoustic models} (AM)
    whose parameters are estimated from parallel data consisting of audio signals and corresponding transcriptions.
    The outcome of the model training depends on the available amount of such parallel data.
    But the amount of available parallel data is usually quite limited, compared to the publicly
    available amount of unimodal data (i.e. audio or textual).
    Consequently, there are attempts to make use of unimodal data to improve the performance of AMs by unsupervised
    pretraining \cite{wav2vec} or combining AMs with external \textit{language models} (LM) which have been trained
    on text only data \cite{GFX+15}.

    Previous work \cite{MSN20} shows that integrating LMs into the training of attention-based encoder-decoder models
    via a log-linear combination followed by a renormalization on sequence-level yields promising improvements similar
    to the maximum mutual information (MMI) training of hybrid models.
    However, since the sum over all possible word sequences used in the renormalization cannot be entirely computed
    in practice, it is usually approximated by a sum over an n-best list obtained from \textit{beam search} (BS) with
    finite beam size.
    The number of sequences which can be aggregated through this approach is usually limited to 2-8 competing hypotheses
    for both attention \cite{minWER-google} and transducer models \cite{minWER-tencent,minWER-HAT}.

    In the hybrid approach, this limitation is alleviated by the use of lattices \cite{hybrid-lattice},
    where hypotheses sharing a common history determined by the LM context length are recombined.
    The lattice approach for recognition has been explored in \cite{attention-lattice} for temporal convolution
    models, but it is not directly applicable for general \textit{attention-based encoder-decoder} (AED)
    models.

    In this work an extension to the n-best list approach is proposed. Introducing recombination operations in the
    standard beam search procedure the n-best list is transformed into a lattice which contains a denser and 
    much richer representation of the search space.
    Using lattices, the number of sequences considered for the approximation of the renormalization can be increased
    significantly.

    \section{Training Criterion}
    \label{sec:globalrenorm}

    In the following, let \(w_1^N := w_1...w_n...w_N\) denote a word sequence of length \(N \in \mathbb{N}\), and let
    \(x_1^T := x_1...x_t...x_T\) denote an input signal sequence of length \(T \in \mathbb{N}\).

    In this work, an attention-based AM \(p_\mathrm{AM}(w_1^N | x_1^T)\) and an external LM \(p_\mathrm{LM}(w_1^N)\) are
    combined via log-linear model combination. The (normalized) probability of the \textit{combined model} is given by
    \begin{align}
    \label{eqn:combined-model}
        p_c(w_1^N | x_1^T) = \frac{p_\mathrm{AM}^\alpha(w_1^N | x_1^T) \cdot p_\mathrm{LM}^\beta (w_1^N)}{\sum_{\tilde{N}, \tilde{w}_1^{\tilde{N}}} p_\mathrm{AM}^\alpha(\tilde{w}_1^{\tilde{N}} | x_1^T) \cdot p_\mathrm{LM}^\beta (\tilde{w}_1^{\tilde{N}})}
    \end{align}
    where the influence of AM and LM on the output of the combined model is scaled by the AM scale
    \(\alpha \in \mathbb{R}^+\) and the LM scale \(\beta \in \mathbb{R}^+\).
    The sum in the denominator contains all possible word sequences and is infeasible to compute in practice.
    For recognition, the normalization term is not needed, as it does not influence the result. In training, however, 
    the sum does appear in the training criterion
    \begin{equation}
    F = \log \, p_c(w_1^N | x_1^T).
    \label{eqn:training_criterion}
    \end{equation}

    \subsection{Approximative Lattice Recombination}
    \label{sec:bsr}
    
    In the hybrid approach, sequence training can be performed efficiently with the help of lattices. Here, the acoustic
    and language model only take a limited number $k$ of past words into account. So if the last $k$ words of two
    hypotheses match they can be recombined exactly.
    
    For attention models or transducers with unlimited history context, this can never happen. It is possible to 
    force a recombination with different label histories $w_1^n$ and $\tilde{w}_1^n$ under the approximation that for 
    every following step $m$ 
    \begin{equation}
        p_c(w | w_1^m, x_1^T) \approx  p_c(w | w_{n+1}^m, \tilde{w}_1^n, x_1^T)
        \label{eqn:approx}
    \end{equation}
    holds. In the following the proposed procedure to exploit this approximation during beam search is described.
    
    Let the \textit{recombination history limit} $k \in \mathbb{N}$ be the minimum number of tokens of common history
    after which the approximation of \autoref{eqn:approx} is assumed to be sufficiently well met. Then let 
    \(B_n(w_{n-k}^{n-1})\) denote the set of sequences within the beam at decoding step \(n\), which share the common
    history \(w_{n-k}^{n-1}\).

    Before the hypotheses are expanded with the predictions for step \(n\), candidates for recombination are identified.
    For each set of hypotheses that can be recombined, the sequence with the highest score so far is selected
    \begin{equation}
     \hat{w}_1^{n-1} = \argmax_{w_1^{n-1} \in B_n(w_{n-k}^{n-1})} p(w_1^{n-1}|x_1^T)
    \end{equation}
    and the score mass of all other hypotheses in \(B_n(w_{n-k}^{n-1})\) is combined into the hypothesis of 
    \(\hat{w}_1^{n-1}\). Then the other sequences are removed from the search space and the hypothesis of 
    \(\hat{w}_1^{n-1}\) now holds the probability mass 
    \begin{equation}
    \hat{p}(\hat{w}_1^{n-1}|x_1^T) = \sum_{w_1^{n-1} \in B_n(w_{n-k}^{n-1})} p(w_1^{n-1}|x_1^T).
    \end{equation}
    Subsequently, the branch expansion continues with the remaining hypotheses and altered scores.
    Once all sequences in the beam are terminated with a sentence-end token, the search terminates and the probability
    mass is summed to approximate the sum of the scores of all possible word sequences.

\begin{figure}
        \ifRenderfigures
        \centering
        \resizebox{\columnwidth}{!}{
            \def \unitsep{2cm}
\def \unitsepx{2*\unitsep}
\def \unitsepy{\unitsep}

\begin{tikzpicture}  
    \Large

    \tikzstyle{point}=[draw,circle,very thick,fill=rwthlightblue,minimum width=16pt,inner sep=0pt];
    \tikzstyle{point master}=[point,fill=rwthorange];

    \def \radius{0.2}
    \def \arrwidth{5pt}
    
    \def \x{2}
    \def \yone{2}
    \def \ytwo{3}
    \draw[ultra thick,rounded corners=12pt,fill=rwthorange] (\unitsepx*\x-\radius*\unitsep,\yone*\unitsepy-\radius*\unitsep) -- (\unitsepx*\x+\radius*\unitsep,\yone*\unitsepy-\radius*\unitsep) -- (\unitsepx*\x+\radius*\unitsep,\ytwo*\unitsepy+\radius*\unitsep) -- (\unitsepx*\x-\radius*\unitsep,\ytwo*\unitsepy+\radius*\unitsep) -- cycle;
    \draw[line width=\arrwidth,-{Triangle[width=2*\arrwidth,length=1.8*\arrwidth]}](\unitsepx*\x, \yone*\unitsepy+0.3*\unitsep) -- (\unitsepx*\x, \yone*\unitsepy+0.7*\unitsep);

    \def \x{3}
    \def \yone{2}
    \def \ytwo{3}
    \draw[ultra thick,rounded corners=12pt,fill=rwthorange] (\unitsepx*\x-\radius*\unitsep,\yone*\unitsepy-\radius*\unitsep) -- (\unitsepx*\x+\radius*\unitsep,\yone*\unitsepy-\radius*\unitsep) -- (\unitsepx*\x+\radius*\unitsep,\ytwo*\unitsepy+\radius*\unitsep) -- (\unitsepx*\x-\radius*\unitsep,\ytwo*\unitsepy+\radius*\unitsep) -- cycle;
    \draw[line width=\arrwidth,-{Triangle[width=2*\arrwidth,length=1.8*\arrwidth]}](\unitsepx*\x, \yone*\unitsepy+0.7*\unitsep) -- (\unitsepx*\x, \yone*\unitsepy+0.3*\unitsep);

    \def \seqlength{4}
    \def \beamsize{4}
    
    \node[point] (origin) at (0,2.5*\unitsep) {};
    
    \foreach[evaluate={\xminone=\x-1}] \x in {1,...,\seqlength}
    {
        \foreach[evaluate={\yminone=\y-1}] \y in {1,...,\beamsize}
        {   
            \ifthenelse{\(\x=2 \AND \y=3\) \OR \(\x=3 \AND \y=2\)} 
            {
                \node[point master] (p-\x-\y) at (\x*\unitsepx,\y*\unitsepy) {};
            }{
                \node[point] (p-\x-\y) at (\x*\unitsepx,\y*\unitsepy) {};
            }
            
        }
    }
    
    \path (origin) edge[ultra thick] node[above,pos=0.5]{D} (p-1-1);
    \path (origin) edge[ultra thick] node[above,pos=0.5]{C} (p-1-2);
    \path (origin) edge[ultra thick] node[above,pos=0.5]{B} (p-1-3);
    \path (origin) edge[ultra thick] node[above,pos=0.5]{A} (p-1-4);
    
    \path (p-1-1) edge[ultra thick] node[above,pos=0.5]{C} (p-2-1);
    \path (p-1-2) edge[ultra thick] node[above,pos=0.5,text=rwthorange]{\textbf{B}} (p-2-2);
    \path (p-1-3) edge[ultra thick] node[above,pos=0.5,text=rwthorange]{\textbf{B}} (p-2-3);
    \path (p-1-3) edge[ultra thick] node[above,pos=0.5]{A} (p-2-4);
    
    \path (p-2-3) edge[ultra thick] node[above,pos=0.5]{C} (p-3-1);
    \path (p-2-3) edge[ultra thick] node[above,pos=0.5,text=rwthorange]{\textbf{D}} (p-3-2);
    \path (p-2-4) edge[ultra thick] node[above,pos=0.5,text=rwthorange]{\textbf{D}} (p-3-3);
    \path (p-2-4) edge[ultra thick] node[above,pos=0.5]{A} (p-3-4);
    
    \path (p-3-2) edge[ultra thick] node[above,pos=0.5]{D} (p-4-1);
    \path (p-3-2) edge[ultra thick] node[above,pos=0.5]{C} (p-4-2);
    \path (p-3-4) edge[ultra thick] node[above,pos=0.5]{B} (p-4-3);
    \path (p-3-4) edge[ultra thick] node[above,pos=0.5]{A} (p-4-4);
    
    \node[right of=p-4-4,anchor=west] {BAAA};
    \node[right of=p-4-3,anchor=west] {BAAB};
    \node[right of=p-4-2,anchor=west] {BA\textcolor{rwthorange}{D}C, B\textcolor{rwthorange}{BD}C, C\textcolor{rwthorange}{BD}C};
    \node[right of=p-4-1,anchor=west] {BA\textcolor{rwthorange}{D}D, B\textcolor{rwthorange}{BD}D, C\textcolor{rwthorange}{BD}D};

\end{tikzpicture}
        }
        \else
        {\centering{[Renderfigures is disabled]}}
        \fi
        \caption[Example: BSR trellis]{
            Illustration of lattice recombination with beam size \(b=4\) and recombination history limit \(k=1\).
            The search progresses from left to right.
            The matching histories are highlighted in \textcolor{rwthorange}{orange}.
            The sequences considered by the search are listed on the right.
        }
        \label{fig:bsr-example}
    \end{figure}
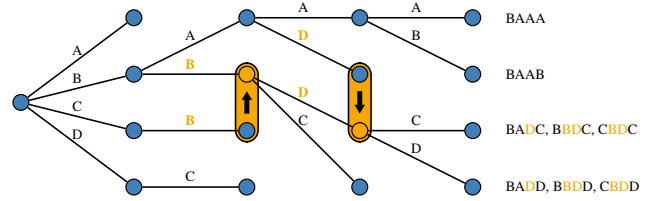

    An exemplary lattice built by this search procedure together with the sequences it contains is illustrated in
    \autoref{fig:bsr-example}.

\pagebreak
\vspace{-3mm}
    \section{Experimental setup}
    \label{sec:setup}
    \vspace{-2mm}
    All experiments use the Sisyphus~\cite{PBN18} workflow manager and RETURNN~\cite{DZV+17}. 
    The configuration files of the experiments are available online.
    \footnote{\url{https://github.com/rwth-i6/returnn-experiments/tree/master/2022-approx-recombination}}

    The experiments of this work are conducted on the Librispeech task and use two kinds of models, one attention-based encoder-decoder model
    with a decoder with limited label context length and one with unlimited label context length.
    In both cases, the same encoder architecture is used for the AM, which consists of 2 CNN layers, followed by 6 BLSTM
    layers which perform time sub-sampling by a factor of 6 via max-pooling.
    As input, the encoder receives 40 dimensional MFCC features which are perturbed by a variant of
    SpecAugment~\cite{PCZ+19}.
    Further, in both cases, MLP-style attention with weight feedback is used.
    Note that the weight feedback adds some recurrence to the decoder of the AM with limited context length.
    The output vocabulary consists of 10k grapheme BPE tokens.

    For the AMs with unlimited context length, the decoder is based on a single LSTM layer with 1000 units
    \cite{MSN20,ZBI+19,Zeineldeen-ILM}.    
    For the AMs with limited context length, the decoder consists of a single FFNN layer, which receives an attention
    vector as well as the last 5 target embeddings as input \cite{Zeineldeen-ILM}.

    For the construction of combined models, the LSTM-decoder based AM is combined with an LSTM LM consisting of a 
    single LSTM layer and the FFNN AM is combined with a FFNN LM which consists of 3 FFNN layers and also has a 
    label context length of 5.
    Both LMs were trained on the audio transcriptions only.
    While an LSTM LM or an FFNN LM respectively is used for training, a Transformer LM consisting of 24 layers and 
    trained on additional text data is used during decoding in any case.
    The LMs operate on the same vocabulary as the acoustic models do.

    AMs and LMs are first trained separately with the \textit{cross-entropy} (CE) criterion until they converge, 
    before they are used to initialize the combined models. The AM training includes a pre-training scheme with 
    gradually increasing layer sizes and an additional CTC-loss on the encoder outputs \cite{ZBI+19}. 
    The combined model is then fine tuned with the sequence discriminative objective function in
    \autoref{eqn:training_criterion} using different recombination history limits \(k\).
    During the training of the combined models, only the parameters of the underlying AMs are updated.

    In training, an AM scale of \(\alpha = 0.1\) and an LM scale of \(\beta=0.035\) is used, which were found to be
    optimal in \cite{MSN20}.
    For decoding the scales are individually tuned on the respective dev sets.

    \begin{table}[tb]
        \caption{
            Comparison of lattices statistics obtained with different recombination history limits \(k\) by averaging 
            searches over the first 1024 segments of the LibriSpeech dev-other set.
            The rows where \(k=\infty\) correspond to n-best lists.
        }
        \label{tab:bsr-stats}
        \begin{subtable}{\columnwidth}
            \caption{Combined model with unlimited context length.}
            \label{tab:bsr-stats-lstm}
            \centering
            \begin{tabular}{c||c|c|c|c}
    \toprule
	history       & log-score     & \multicolumn{2}{c|}{number of}                          & Euclid.  \\
	limit \(k\)   & mass          & sequences                         & recomb.             & dist.    \\ \midrule
	1             & \(-7.098\)    & \(1.2 \cdot 10^{13}\)             & \(15.8\)            & \(0.195\)\\
    2             & \(-8.032\)    & \(1.2 \cdot 10^{7\phantom{0}}\)   & \(10.0\)            & \(0.106\)\\
    3             & \(-8.522\)    & \(3.1 \cdot 10^{3\phantom{0}}\)   & \(\phantom{0}7.5\)  & \(0.077\)\\
    4             & \(-8.777\)    & \(1.9 \cdot 10^{3\phantom{0}}\)   & \(\phantom{0}6.0\)  & \(0.060\)\\
    6             & \(-9.166\)    & \(4.3 \cdot 10^{1\phantom{0}}\)   & \(\phantom{0}4.1\)  & \(0.041\)\\
    8             & \(-9.299\)    & \(2.3 \cdot 10^{1\phantom{0}}\)   & \(\phantom{0}3.1\)  & \(0.028\)\\
    10            & \(-9.466\)    & \(1.7 \cdot 10^{1\phantom{0}}\)   & \(\phantom{0}1.0\)  & \(0.023\)\\
    12            & \(-9.560\)    & \(1.4 \cdot 10^{1\phantom{0}}\)   & \(\phantom{0}1.9\)  & \(0.018\)\\ \midrule
	$\infty$      & \(-9.907\)    & \(0.8 \cdot 10^{1\phantom{0}}\)   & \(\phantom{0}0.0\)  & \(-\)    \\
    \bottomrule
\end{tabular}


        \end{subtable}

        \vspace{2ex}
        \begin{subtable}{\columnwidth}
            \caption{Combined model with limited context length of \(5\).}
            \label{tab:bsr-stats-ffnn}
            \centering
            \begin{tabular}{c||c|c|c|c}
    \toprule
	history       & log-score       & \multicolumn{2}{c|}{number of}                            & Euclid.       \\
	limit \(k\)   & mass            & sequences                         & recomb.               & dist.         \\ \midrule
	1             & \(-6.836\)      & \(7.2 \cdot 10^{12}\)             & \(15.9\)              & \(0.2392\)    \\
    2             & \(-7.824\)      & \(3.5 \cdot 10^{7\phantom{0}}\)   & \(\phantom{0}9.9\)    & \(0.1363\)    \\
    4             & \(-8.639\)      & \(2.8 \cdot 10^{3\phantom{0}}\)   & \(\phantom{0}5.9\)    & \(0.0842\)    \\
    5             & \(-8.832\)      & \(6.3 \cdot 10^{2\phantom{0}}\)   & \(\phantom{0}4.8\)    & \(0.0028\)    \\
    8             & \(-9.270\)      & \(2.7 \cdot 10^{1\phantom{0}}\)   & \(\phantom{0}3.0\)    & \(0.0002\)    \\ \midrule
	$\infty$      & \(-9.879\)      & \(0.8 \cdot 10^{1\phantom{0}}\)   & \(\phantom{0}0.0\)    & \(-\)         \\
    \bottomrule
\end{tabular}


        \end{subtable}
    \end{table}

    \section{Results}
    \label{sec:results}

    \subsection{Lattice Analysis}
    \label{sec:bsranalysis}

    To investigate how the lattice approximation behaves in practice compared to the standard n-best list, 
    lattice statistics are extracted for the combined model both with unlimited and with limited context length of \(5\).
    In both cases, the combined model just after initialization with the respective underlying AM and LM is used.

    For the experiment, beam search is conducted over each of the first 1024 segments of the 
    LibriSpeech dev-other set for different recombination history limits \(k\).
    Additionally, a beam search without recombination is conducted for comparison, which in the following is
    reported as \(k=\infty\). The beam size for each search is set to \(b=8\).

    For each search the number of word sequences that can be obtained from the final lattice, the amount of 
    aggregated score mass, and the number of recombinations are measured.
    On each recombination, the context with the highest sequence score so far is selected to replace the 
    contexts of the other recombined branches. In future steps the correct probability distribution of these branches 
    is displaced by an approximative distribution with the updated context.
    The difference between these two distributions is estimated by averaging the Euclidean distance between the
    best-context distribution and each recombined distribution.
    \begin{equation}
    d = \sum_w \left| p_n(w|\hat{w}_1^{n-1}, x_1^T) - p_n(w|w_1^{n-1}, x_1^T) \right|^2
    \end{equation}

    The statistics obtained from each segment search are averaged for each \(k\) and displayed in \autoref{tab:bsr-stats}.

    As to be expected, lowering \(k\) increases the number of recombinations which are made during search, which in turn
    increases the number of sequences considered and the score mass aggregated through a search.
    In case of single word recombination history limit compared to the case without recombination \((k=\infty)\), 
    the average number of considered sequences
    is increased by a factor of \(10^{12}\) and a \(20\)-fold greater average score mass can be aggregated.
    However, lowering \(k\) also worsens the approximations, as seen by the increasing Euclidean distances.

    For the limited context model, the Euclidean distance lies above the the distance measured for the unlimited 
    context model for \(k<5\). As the recombination history limit reaches the model context length, a significant 
    drop in the 
    distance is observed and it is now more than one order of magnitude lower than the comparable distance of the full
    context model. The distance does, however, not immediately reach exactly zero. This is due to the attention feedback
    of the underlying AM which applies some recurrence and hence some dependence on the complete context to the model.

    \subsection{Training results}
    \label{sec:trainingresults}
    
    In the next step, lattice recombination is used to approximate the denominator in \autoref{eqn:combined-model} 
    during training. The trainings are conducted using the beam sizes \(b=8\) and \(b=4\). All tables also
    include the results of using n-best lists for the approximation of the denominator sum in the row \(k=\infty\)
    for comparison.
    
    \vspace{-2mm}
    \paragraph*{Unlimited Context Models}
    For the model with unlimited context length, multiple recombination history limits \(k\) are evaluated.
    The training results are listed in \autoref{tab:bsr-k-lstm}.
    
    It can be seen that in the limit of large \(k\), the result of n-best lists can be recovered. Reducing \(k\)
    degrades the \textit{word error rate} (WER) of the final model on the LibriSpeech dev sets. 
    The benefit of a larger effective beam size is surpassed by the error that is incurred by the stronger approximations.
    
    In case of the models trained with a beam size of \(b=4\), a small degradation compared to \(b=8\) is observed.
    Here, reducing the recombination history limit does not change the results significantly. Even with a higher 
    effective beam size, the \(b=8\) result cannot be recovered.
    
    \vspace{-2mm}
    \paragraph*{Limited Context Models}    
    In case of the combined model with limited context length, the recombination history limit is chosen to be 
    equal to the model context length \(k=5\). 
    The results are listed in \autoref{tab:bsr-k-ffnn}.

    Using lattices with a recombination history limit of \(k=5\) yields small improvements over the use of n-best 
    lists on dev-other and similar results on dev-clean. With the smaller beam size \(b=4\) the same result as with 
    \(b=8\) is achieved but also no clear degradation is visible in the case of n-best lists.
    

    The results from the initial model without sequence training together with the test results of the combined models
    which performed best on the dev sets are listed in \autoref{tab:test} and are consistent with the observations made on the dev sets.

    \begin{table}
        \caption{
            Comparing performances of combined models with unlimited context length trained with different 
            recombination history limits \(k\), where \(k=\infty\) means using n-best list.
        }
        \label{tab:bsr-k-lstm}

        \begin{subtable}[t]{0.49\columnwidth}
            \caption{
                Denom. beam size \(b=8\)
            }
            \label{tab:bsr-k-b8-lstm}
            \centering
            \begin{tabular}{c||c|c}
    \toprule
	history       & \multicolumn{2}{c}{dev WER[\%]} \\
	limit \(k\)   & clean       & other             \\ \midrule
	1             & \(2.2\)     & \(6.2\)           \\
    2             & \(2.1\)     & \(6.2\)           \\
    3             & \(2.1\)     & \(6.1\)           \\
    4             & \(2.1\)     & \(6.0\)           \\
    6             & \(2.1\)     & \(5.9\)           \\
    8             & \(2.1\)     & \(5.9\)           \\
    10            & \(2.1\)     & \(5.8\)           \\
    12            & \(2.1\)     & \(5.8\)           \\ \midrule
	$\infty$      & \(2.1\)     & \(5.8\)           \\
    \bottomrule
\end{tabular}
        \end{subtable}
        \begin{subtable}[t]{0.49\columnwidth}
            \caption{
                Denom. beam size \(b=4\)
            }
            \label{tab:bsr-k-b4-lstm}
            \centering
            \begin{tabular}{c||c|c}
    \toprule
	history       & \multicolumn{2}{c}{dev WER[\%]} \\
	limit \(k\)   & clean       & other             \\ \midrule
	1             & \(2.2\)     & \(6.2\)           \\
    2             & \(2.1\)     & \(6.0\)           \\
    3             & \(2.1\)     & \(6.0\)           \\
    4             & \(2.1\)     & \(5.9\)           \\
    6             & \(2.1\)     & \(6.0\)           \\
    8             & \(2.1\)     & \(6.0\)           \\
    10            & \(2.1\)     & \(5.9\)           \\
    12            & \(2.1\)     & \(6.0\)           \\ \midrule
	$\infty$      & \(2.1\)     & \(6.0\)           \\
    \bottomrule
\end{tabular}
        \end{subtable}
    \end{table}

    \begin{table}
        \caption{
            Comparing performance of combined models with limited context length of \(5\) trained with 
            recombination history limit of \(k=5\) and \(k=\infty\), which corresponds to using n-best list.
        }
        \label{tab:bsr-k-ffnn}

        \begin{subtable}[t]{0.49\columnwidth}
            \caption{
                Denom. beam size \(b=8\)
            }
            \label{tab:bsr-k-b8-ffnn}
            \centering
            \begin{tabular}{c||c|c}
    \toprule
	history       & \multicolumn{2}{c}{dev WER[\%]} \\
	limit \(k\)   & clean       & other             \\ \midrule
    5             & \(2.3\)     & \(6.2\)           \\ \midrule
	$\infty$      & \(2.3\)     & \(6.4\)           \\
    \bottomrule
\end{tabular}
        \end{subtable}
        \begin{subtable}[t]{0.49\columnwidth}
            \caption{
                Denom. beam size \(b=4\)
            }
            \label{tab:bsr-k-b4-ffnn}
            \centering
            \begin{tabular}{c||c|c}
    \toprule
	history       & \multicolumn{2}{c}{dev WER[\%]} \\
	limit \(k\)   & clean       & other             \\ \midrule
    5             & \(2.3\)     & \(6.2\)           \\ \midrule
	$\infty$      & \(2.4\)     & \(6.3\)           \\
    \bottomrule
\end{tabular}
        \end{subtable}
    \end{table}

    \begin{table}
        \caption{
            Comparing the performances of the discriminatively trained combined models which perform best on the 
            dev sets. WER of the model used for initialization (init) is included for reference.
        }
        \label{tab:test}
        \centering
        \begin{tabular}{c|c||c|c|c|c}
    \toprule
	model           & history       & \multicolumn{2}{c|}{dev WER[\%]}  & \multicolumn{2}{c}{test WER[\%]}  \\
	context         & limit \(k\)   & clean         & other             & clean         & other             \\ \midrule
                    & init          & \(2.6\)       & \(6.7\)           & \(3.0\)       & \(7.1\)           \\
    limited         & 5             & \(2.3\)       & \(6.2\)           & \(2.6\)       & \(6.6\)           \\
                    & $\infty$      & \(2.3\)       & \(6.4\)           & \(2.6\)       & \(6.7\)           \\ \midrule
                    & init          & \(2.4\)       & \(6.7\)           & \(2.7\)       & \(7.0\)           \\
    unlimited       & 12            & \(2.1\)       & \(5.8\)           & \(2.3\)       & \(6.5\)           \\
                    & $\infty$      & \(2.1\)       & \(5.8\)           & \(2.3\)       & \(6.5\)           \\
    \bottomrule
\end{tabular}
    \end{table}

\pagebreak
    \section{Conclusion}

    In this paper, \textit{approximative recombination} is used to improve the summation of the scores of all 
    competing word sequences, which is used in sequence discriminative training of attention-based encoder-decoder models.

    Using lattices, the average number of sequences considered for the summation can be increased by a factor of up to
    \(10^{12}\) and an up to \(20\)-fold larger average score mass can be aggregated compared to standard n-best lists
    (cf. \autoref{tab:bsr-stats}).

    Sequence training with both, n-best lists and lattices, show good improvement over the cross entropy baseline model.
    In the limit of large recombination history limits \(k\), the n-best list result can be recovered, but for smaller 
    \(k\) only the limited context FFNN decoder model shows additional improvement over standard beam search.

    Nevertheless, approximative recombination may be a useful tool for applications that benefit from the larger 
    search space coverage without being handicapped by the additional approximation, such as lattice rescoring or
    keyword spotting.

    \section{Acknowledgments}
    \begin{wrapfigure}[4]{l}{0.13\textwidth}
	    \vspace{-8mm}
	    \begin{center}
	    	\includegraphics[width=0.15\textwidth]{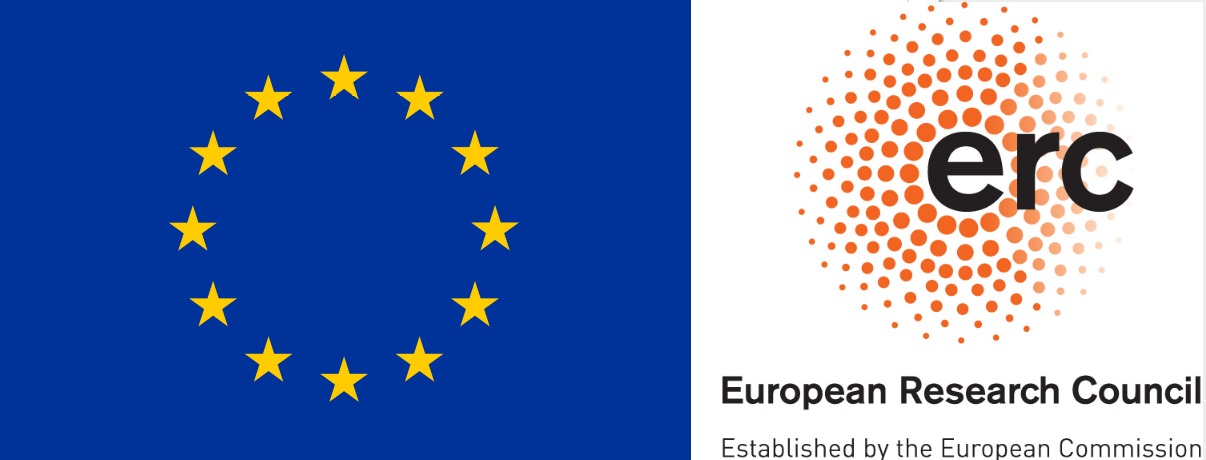} \\
	    \end{center}
	    \vspace{-4mm}
    \end{wrapfigure}
    This work has received funding from the European Research Council (ERC) under the European Union's Horizon 2020
    research and innovation programme (grant agreement No 694537, project "SEQCLAS"). 
    The work reflects only the authors' views and none of the funding parties is responsible for any use that may be 
    made of the information it contains. 

    The authors would like to thank Mohammad Zeineldeen and Aleksandr Glushko for providing the baseline AMs and 
    helpful discussions. 


    \bibliographystyle{IEEEbib}
    \bibliography{literature}

\end{document}